\documentclass[none]{jmlr_arxiv}

\usepackage{longtable}

\usepackage{booktabs}
\usepackage[load-configurations=version-1]{siunitx} 

\usepackage{listings}

\usepackage{hyperref}

\usepackage{authblk}

\makeatletter
\def\set@curr@file#1{\def\@curr@file{#1}} 
\makeatother

\title[Patient Contrastive Learning]{Patient Contrastive Learning: a Performant, Expressive, and Practical Approach to ECG Modeling}
\author[1, 2]{Nathaniel Diamant}
\author[1, 3]{Erik Reinertsen}
\author[3]{Steven Song}
\author[3, 4, 5]{Aaron Aguirre}
\author[1, 3, 6, 7]{Collin Stultz}
\author[2]{Puneet Batra}

\affil[1]{\footnotesize Research Laboratory of Electronics, MIT}
\affil[2]{\footnotesize Data Sciences Platform, Broad Institute of MIT and Harvard}
\affil[3]{\footnotesize Division of Cardiology, Massachusetts General Hospital}
\affil[4]{\footnotesize Center for Systems Biology, Massachusetts General Hospital Research Institute and Harvard Medical School}
\affil[5]{\footnotesize Wellman Center for Photomedicine, Massachusetts General Hospital Research Institute and Harvard Medical School}
\affil[6]{\footnotesize Department of Electrical Engineering and Computer Science, MIT}
\affil[7]{\footnotesize Harvard-MIT Division of Health Sciences and Technology}

\firstpageno{1}
\begin{document}
\maketitle
\vspace{-10mm}
\begin{abstract}
Supervised machine learning applications in health care are often limited due to a scarcity of labeled training data.
To mitigate this effect of small sample size, we introduce a pre-training approach, \textbf{P}atient \textbf{C}ontrastive \textbf{L}earning of \textbf{R}epresentations (PCLR), which creates latent representations of ECGs from a large number of unlabeled examples.  
The resulting representations are expressive, performant, and practical across a wide spectrum of clinical tasks.
We develop PCLR using a large health care system with over 3.2 million 12-lead ECGs, and demonstrate substantial improvements across multiple new tasks when there are fewer than 5,000 labels.
We release our model to extract ECG representations at \url{https://github.com/broadinstitute/ml4h/tree/master/model_zoo/PCLR}.
\end{abstract}

\section{Introduction}
Scarcity of labeled training data prevents the full clinical impact of supervised machine learning in health care. 
As one example, sudden cardiac death (SCD) kills over 450,000 Americans per year \citep{scd}, yet large observational datasets, which contain millions of patient records, typically only have data for a small number of SCDs. 
Unlike machine learning applications outside of health care, it is not routinely possible to significantly increase the number of cases by labeling more data points because the prevalence of the disorder of interest is often very low. 
The resulting lack of statistical power is a significant impediment to the development of accurate risk models \citep{risk}. 

\begin{figure}[htbp]
  \centering 
  \includegraphics[width={0.9 \textwidth}]{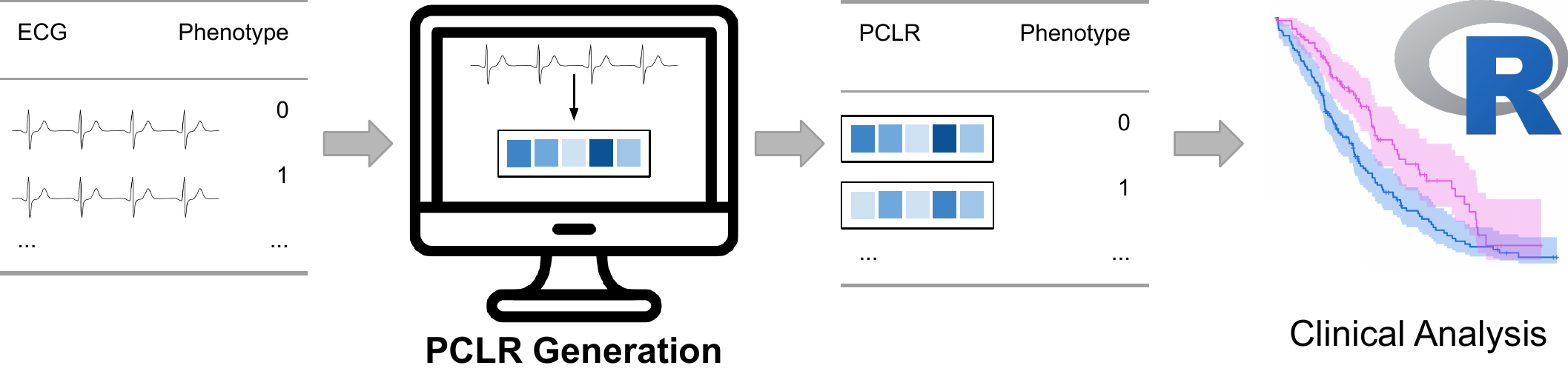}
  \caption{Example workflow using PCLR. No knowledge of deep learning is required.}
  \label{fig:service}
\end{figure}

An approach that has proven successful when there are few labeled training examples is pre-training: neural networks are first trained on a large corpus of data for a set of related tasks, then the pre-trained model is fine-tuned on the task of interest. 
Pre-training has proven successful across many domains, including health care \citep{big_self_sup}.  
The ideal pre-training strategy is: 
\begin{itemize}
\item Performant: it maximizes performance on limited training data)
\item Expressive: it can be used to develop models for multiple tasks)
\item Practical: it is easy-to-use for those unfamiliar with deep learning)
\end{itemize}
In the past,  pre-training strategies in health-care have  focused on the first goal, with a few notable exceptions that also consider the third goal \citep{clocs}. 
As larger groups of clinical researchers adopt machine learning approaches to more tasks, expressivity and ease-of-use have become more critical. Clinical scientists often do not have the resources, or expertise, needed to retrain deep learning models for their specific task, although they do have the greatest insight into model deployment needs.
We therefore develop and validate Patient Contrastive Learning of Representations (PCLR) for 12-lead ECGs to satisfy all these objectives.  
We demonstrate the effectiveness of PCLR using two large hospital system data sets over four distinct ECG tasks.
PCLR yields feature vectors optimized for {\it linear} models, making them easily used by clinical researchers out-of-the-box (Figure \ref{fig:service}).

How does PCLR do this? PCLR is a uniquely clinically-driven modification of SimCLR \citep{simclr}: a contrastive learning approach that builds expressive representations of high-resolution data. 
However, where SimCLR is graded on whether a model can resolve artificial data augmentations, PCLR is graded on whether it can resolve patient identity over time. 
PCLR maps ECGs acquired at different times from a given patient to the same region within a contrastive latent space.
The success of PCLR compared to disease-specific approaches (see Results) shows that patient-centric latent representations are a new direction for deep learning research deserving attention.
As Sir William Osler said, “The good physician treats the disease; the great physician treats the patient who has the disease.” 
PCLR is a step for deep learning towards this dictum.

\subsection*{Generalizable Insights about Machine Learning in the Context of Health care}
\begin{itemize}
    \item Pre-training is needed to overcome the lack of sufficient training data for deep learning approaches in clinical datasets
    \item PCLR pre-training is a generalizable and performant approach to supervised learning on health-care data with limited training data
    \item Practical approaches like PCLR can enable all clinical researchers, regardless of resources, to accelerate clinical impact using deep-learning.
\end{itemize}

\section{Related Work}
\subsection*{Deep learning on ECGs}
There is a growing body of work that applies techniques from deep image classification to 12-lead ECGs in the presence of large labeled datasets.
\cite{ribeiro} train a residual network in a cohort containing millions of labeled ECGs to classify cardiac blocks and other arrhythmias with high accuracy, and we use their model as a baseline for comparison. 
Residual networks have also been shown to outperform automatic labeling systems \citep{outperform_auto_label}, and even physician labels \citep{ng}, and to triage patients \citep{triage}. 
Latent features of the ECG have also been shown to be useful for a wide range of tasks, such as to regress age from the ECG as a marker of cardiac health \citep{attia_age_sex}, or to predict incident atrial fibrillation (AF) \citep{future_af}, or one-year mortality \citep{mortality}.
We contribute by reducing the need for labels, and by focusing on extracting generally expressive representations from ECGs rather than a representation optimized for a single task.

\subsection*{Contrastive learning}
Contrastive learning is a self-supervised learning method that requires training data only to be labeled with notions of positive pairs (data that go together) and negative pairs (data that are distinct).
The SimCLR procedure introduced by \cite{simclr} shows that contrastive learning yields reduced representations of data that are linearly reusable in new tasks.
Many papers have recently experimented with the SimCLR procedure in the medical domain.
\cite{big_self_sup} used the SimCLR procedure in dermatology and X-ray classification tasks.
They defined positive pairs as both modified versions of the same image and images from different views of the same skin condition.
\cite{mri_contrastive} experimented with SimCLR using many different definitions of a positive pair, including MRI images from different patients that show the same area of the body.
\cite{text_image_contrastive} defines positive pairs across modalities, between an X-ray image and its associated text report.
\cite{clocs} apply the SimCLR procedure to 12-lead ECGs, defining positive pairs by different leads from the same ECG or as different non-overlapping time segments within a single ECG.
They show improved performance in rhythm classification tasks compared to other pre-training strategies in both transfer learning and representation learning regimes.
\cite{spine_temporal_siamese} apply a different contrastive learning procedure introduced by \cite{face_contrastive} in lumbar MRIs, but notably define positive pairs in the same way that PCLR does---as pairs of MRIs from the same patient at different times.
\cite{face_contrastive} also make use of image domain specific data augmentations, such as random rotations.
We build on this work by demonstrating that at a large enough scale, a contrastive loss based on patient identity across different ECGs across time is highly performant, expressive, and practical.

\section{Methods}

\subsection*{PCLR Pre-training}
We use a deep residual convolutional neural network to build representations of ECGs.
In pre-training, the network learns to build representations of ECGs specific to an individual and is therefore rewarded when representations of different ECGs from the same person are similar.
The network is penalized when representations of ECGs from different people are similar.
For example, if patient $A$ has ECG $x_i$ taken in 1988 and ECG $x_j$ taken in 2001, then $(x_i, x_j)$ is a positive pair.
If patient $B$ has ECG $x_k$, then both $(x_i, x_k)$ and $(x_j, x_k)$ are negative pairs.

\begin{figure}[ht]
  \centering
    \includegraphics[width=.6\textwidth]{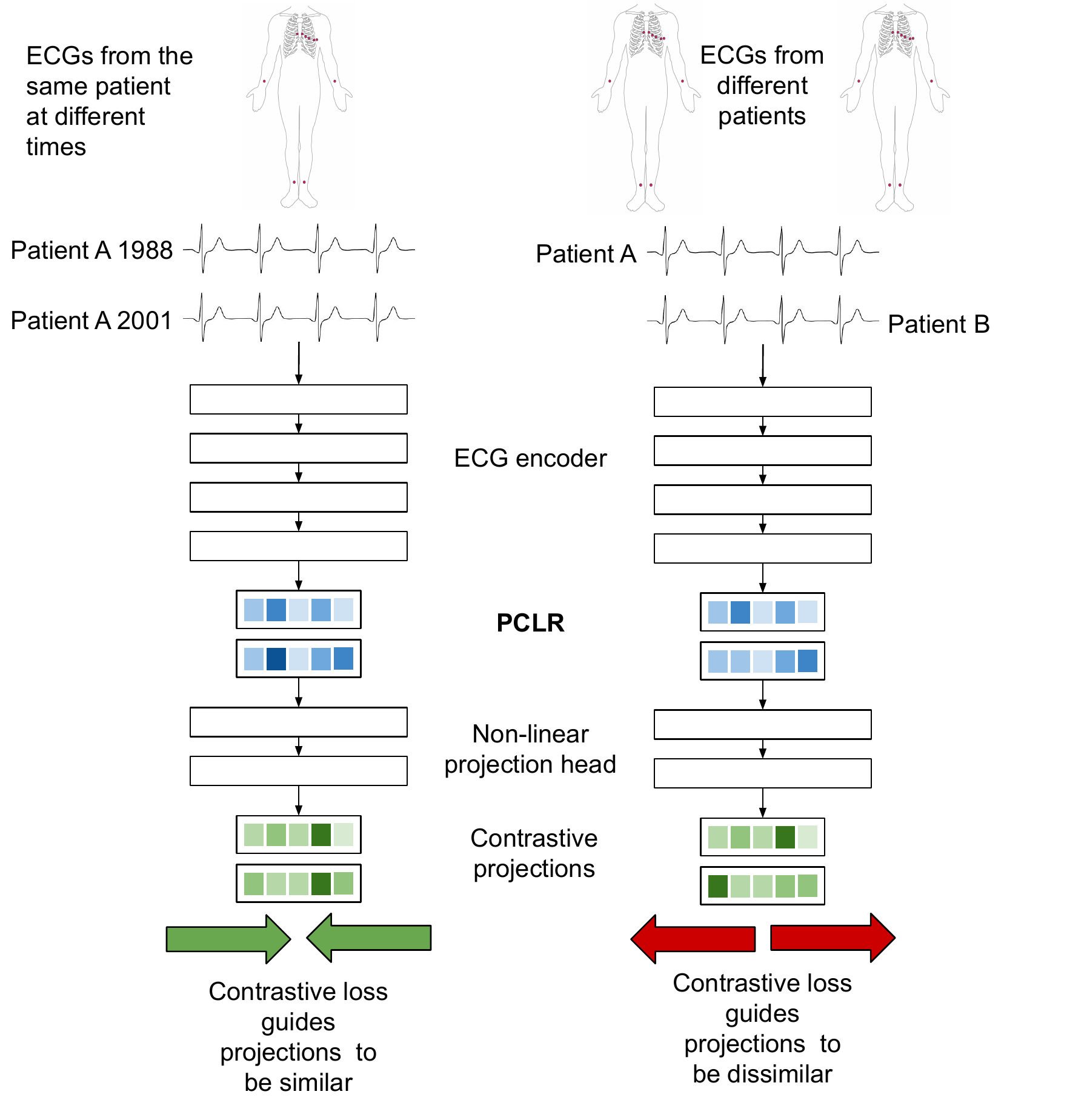}
  \vspace{-5mm}
  \caption{PCLR pre-training procedure for a pair of ECGs from the same patient (left) and from different patients (right).}
  \label{fig:picl_procedure}
\end{figure}

PCLR pre-training has four components (Figure \ref{fig:picl_procedure}):
\begin{enumerate}
    \item An \textit{ECG selection module} which selects pairs of ECGs from individuals.
    For example, it could select ECG $x_i$ from an individual in 1988, and ECG $x_j$ in 2001 from the same participant.
    \item An \textit{ECG encoder} $f(\cdot)$, which produces a compact representation of each 12-lead ECG.
    Given ECG $x_i$, it outputs the encoding $f(x_i) = h_i$. When used in linear models, we refer to this representation as PCLR.
    \item A \textit{projection head} $g(\cdot)$, which projects ECG representations into the space where the contrastive loss is applied.
    For example, $g$ could be applied to $h_i$, giving the projection $z_i = g(h_i)$. \cite{simclr} showed that pre-training with a non-linear projection head improves the usefulness of the learned representation.
    \item A \textit{contrastive loss function} which is used to train the ECG encoder and projection head.
    The contrastive loss function $\ell_{i,j}$ is low when the cosine similarity is high between projections of ECGs from the same patient, $z_i$ and $z_j$.
    $\ell_{i,j}$ also encourages the cosine similarity to be low between ECGs coming from different patients, e.g. $z_i$ and $z_k$.
\end{enumerate}

\subsubsection*{ECG selection module}
Random pairs of ECGs from each individual are selected in every batch (Appendix B) regardless of time or changed health between the ECGs' acquisitions.
We find the random selection approach effective, and leave more advanced strategies to future work.

\subsubsection*{ECG Encoder}
\begin{figure}[htbp]
  \centering 
  \includegraphics[width=.8\textwidth]{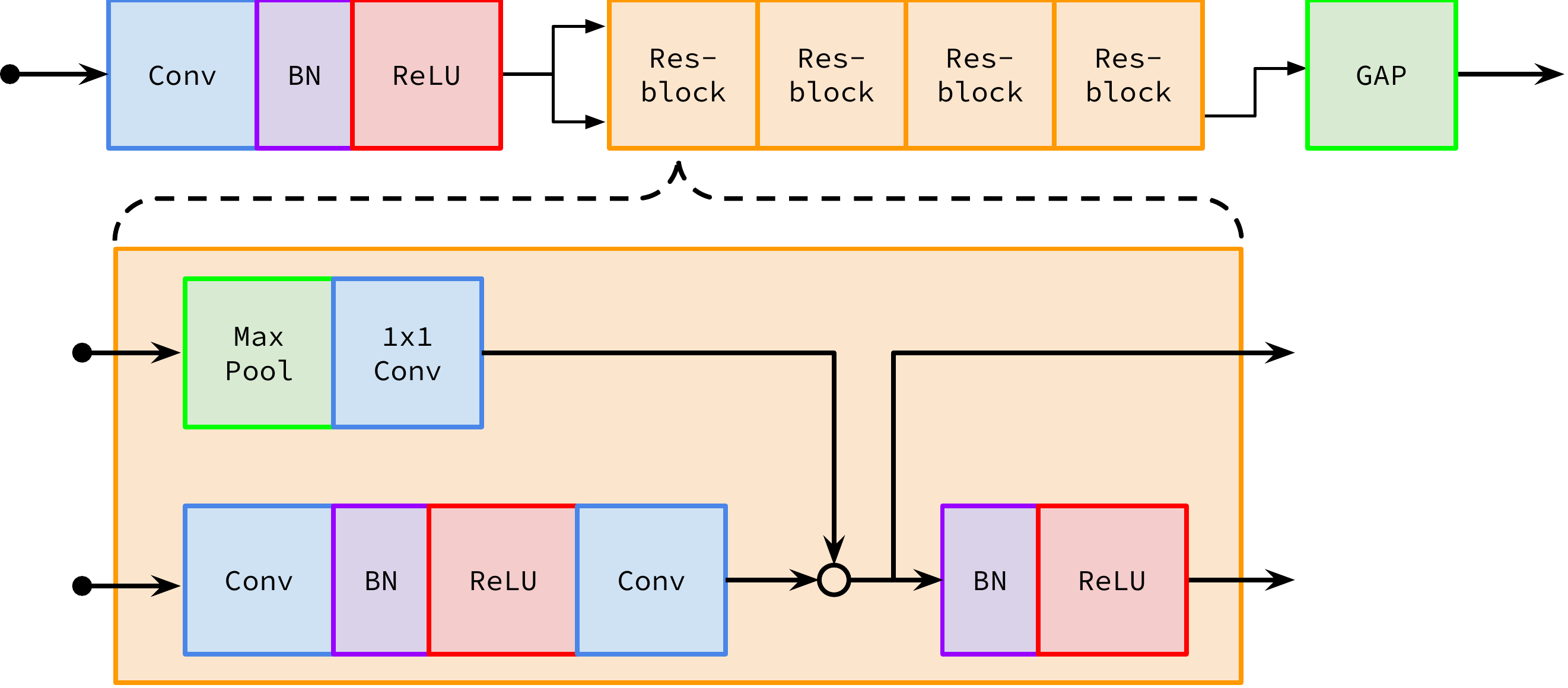} 
  \caption{The ECG encoder architecture designed by \cite{ribeiro} with the addition of global average pooling (GAP). BN is batch normalization.}
  \label{fig:encoder} 
\end{figure}

To facilitate our comparisons, we use the same encoder architecture developed by \cite{ribeiro} (Figure \ref{fig:encoder}). 
We note that our approach will work with any encoding architecture, not just that of \cite{ribeiro}.
In order to adapt it to representation learning, one dimensional global average pooling (GAP) \citep{gap} is applied to the output of the final residual block in the Ribeiro architecture, yielding a 320-dimensional representation for each ECG.

\subsubsection*{Projection Head}
\begin{figure}[ht]
  \centering 
  \includegraphics[width=\textwidth]{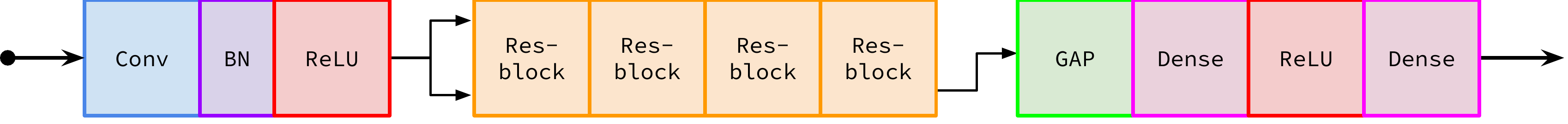} 
  \caption{The full architecture used in PCLR pre-training. The projection head (beginning at the first Dense layer) is applied to the output of the ECG encoder.}
  \label{fig:pretraining_architecture} 
\end{figure}

The projection head follows the encoder and is solely used in pre-training.
\cite{simclr} showed that a non-linear projection head improves the quality of the representations from the encoder.
Our projection head is fully connected layer with 320 units followed by a ReLU activation followed by another 320 unit fully connected layer (Figure \ref{fig:pretraining_architecture}).
The full tensorflow model summary is in Listing \ref{lst:model}.
\subsubsection*{Contrastive loss function}
The contrastive loss function guides the outputs of the projection head to be similar for ECGs that come from the same patient and distinct for ECGs from different patients.
We use the normalized temperature-scaled cross entropy loss with temperature parameter $\tau = 0.1$.
For a pair of ECGs from the same patient, $(x_i, x_j)$, a positive pair, the encoder followed by the projection head yield projections $g(f(x_i)) = z_i$ and $g(f(x_j)) = z_j$.
Given a batch of $N$ patients and letting $\textit{sim}$ denote the cosine similarity yields loss
\begin{equation*}
    \ell_{i,j} = -\log \frac{\exp{[\text{sim}(z_i, z_j) / \tau]}}{\sum_{k=1}^{2N, k \neq i} \exp{[\text{sim}(z_i, z_k)/\tau}]}
\end{equation*}
In a single minibatch of $N$ patients, the loss is computed between all positive pairs.
Let $p_1$ be the index of patient $p$'s first ECG in a minibatch, and $p_2$ be the index of patient $p$'s second ECG in the minibatch.
Then the loss for the batch is
\begin{equation*}
    L_{\text{batch}} = \sum_{p=1}^{N} \ell_{p_1, p_2} 
\end{equation*}

\subsubsection*{PCLR pre-training optimizer details}
During pre-training the mini batch size is 1,024 ECGs drawn from 512 patients.
The model is trained for 50 epochs using the Adam optimizer \citep{adam}.
The learning rate starts at 0.1 and is decayed every epoch according to a half-period cosine schedule with period 50 epochs \citep{cosinedecay}.

\subsection{Applying PCLR to a new task}\label{sec:lin_eval}
We refer to the encoded representations $f(x_i)$ from ECGs as PCLR.
Training linear models on learned representations is known as \textit{linear evaluation}.
Linear evaluation has been shown to be a useful indicator of the performance of more complex models trained on learned representations \citep{linear_eval}.
Furthermore, linear evaluation allows us to test the usefulness of PCLR to practitioners who lack the resources or expertise to train their own neural network and to facilitate integration into existing clinical models (e.g. cox proportional hazard analyses).
We apply linear evaluation to PCLR in a similar procedure to SimCLR:
\begin{enumerate}
    \item Apply the PCLR-trained ECG encoder to $N$ ECGs to get 320 features for each ECG.
    This results in an $N \times 320$ feature vector, which we call PCLR.
    We then normalize each column by subtracting its mean and dividing by its standard deviation.
    \item Train a linear or logistic ridge regression model on PCLR. 
    We use four fold cross validation to select the optimal $\ell^2$ penalty from 10 values logarithimcally-evenly spaced between $10^{-6}$ and $10^5$.
    \item Evaluate the linear model on holdout data.
    We apply the ECG encoder to the hold out ECGs, and normalize the resultant representations using the training summary statistics.
\end{enumerate}

\section{Cohort}
We use ECGs and metadata from two hospitals: Massachusetts General Hospital (MGH) and Brigham and Women's Hospital (BWH).
The data for both cohorts was approved by the Institutional Review Board (IRB) at both hospitals, with a waiver of informed consent.
The data were ingested using the ML4H repository \citep{ml4h}.
We pre-train the ECG encoder using PCLR in a pre-training cohort taken from MGH. 
In the pre-training cohort, we extract age, sex, and ECG sampling rate from each ECG.
The pre-training cohort is further described in Section \ref{sec:pretraining_cohort}.

We build two cohorts from BWH: a set for training supervised models from scratch, and a validation set.
From BWH we extract age, sex, and automatically recorded ECG features: ECG sampling rate, heart rate, PR interval, QRS duration, QT interval, P-axis, R-axis, and T-axis. We also label two disease states from each ECG using a diagnosis text field: atrial fibrillation (AF) and left ventricular hypertrophy (LVH).
AF is an arrhythmia caused by a lack of coordination of the upper and lower chambers of the heart, and is marked on the ECG by an irregular rhythm and missing P-wave.
AF has been associated with a greatly increased risk of stroke \citep{framingham}.
LVH is defined by an increase in left ventricular mass and/or an increase in the left ventricular cavity.
LVH can be detected from the ECG using voltage and interval criteria with high specificity but low sensitivity \citep{cornell_lvh}.
LVH is associated with an increased risk of cardiovascular diseases and death \citep{lvh}.
The two BWH datasets are further described in Section (\ref{section:test_cohort}).

\subsection{Data Extraction}
In both the pre-training and test cohorts, age, sex, heart rate, PR interval, QRS duration, and QT interval are all reported as tabular fields.
LVH and AF come from a free text diagnosis field by checking for containment of keywords (Appendix C).
The ECG waveforms in both cohorts are recorded for ten seconds at either 250 Hz or 500 Hz with amplitudes represented in microvolts as 16 bit integers.
All 12 leads of the ECG are recorded in their own fields.
As preparation for training the ECG encoder, we divide the amplitudes for each lead by 1,000 to get units of millivolts, convert to 32 bit floats, and then use linear interpolation to fit each lead into 4,096 samples.
Once each lead has been interpolated to the same length, we them into a $4,096 \times 12$ matrix with lead order \{I, II, III, AVR, AVL, AVF, V1, V2, V3, V4, V5, V6\}.
The ECG pre-processing code is available at \url{https://github.com/broadinstitute/ml4h/tree/master/model_zoo/PCLR}.

\subsection{Pre-training cohort from MGH}\label{sec:pretraining_cohort}

We filter out patients with only one ECG in order to make the PCLR contrastive loss more informative.
That leaves 404,929 patients with 3,229,408 ECGs.
We randomly select 90\% of the patients into a training set and place the remaining 10\% of patients in the validation set.
The summary statistics of the pre-training cohort are shown in Table \ref{tab:pretrain}.
The validation set is used to pick the checkpoint of the model with the best performance during training.

\begin{table}[htbp]
    \centering
    \caption{PCLR pre-training training and validation sets from MGH}
    \label{tab:pretrain}
    \begin{tabular}{lrr}
    \toprule
    & training & validation \\
    \midrule
        Total Patients & 364,436 & 40,493 \\
        Total ECGs & 2,907,705 & 321,703 \\
        ECGs/patient (median; mean) & 4.0; 7.98 $\pm$ 11.7 & 4.0; 
        7.94 $\pm$ 11.7 \\
        Time between ECGs days (median; mean) & 76; 262 $\pm$ 467 & 75; 261 $\pm$ 467 \\
        Age (median; mean) & 64.0; 62.2 $\pm$ 16.9 & 64.0; 62.1 $\pm$ 17.1 \\
        Sex (Male : Female) & 192,072 : 169,633 & 21,278 : 18,894 \\
        Sampling rate (250Hz : 500 Hz) & 1,939,279 : 968,426 & 214,476 : 107,227 \\
    \bottomrule
    \end{tabular}
\end{table}

\subsection{BWH cohorts} \label{section:test_cohort}
We produce a test set from BWH (BWH-test) with 10,000 ECGs, and a series of training sets of increasing size:
640, 1280, 2560, 5120, 10240, and 20480.
We call these BWH-640, BWH-1280, etc.
Each of the BWH-series is a subset of the one larger than it.
For example, BWH-1280 is a subset of BWH-2560.
The selection criteria for ECGs in the BWH-series and BWH-test are:
\begin{enumerate}
    \item Require that the patient has no ECGs recorded in MGH
    \item Take only the most recent ECG from each patient
    \item Restrict age to between 20 and 90 years
    \item Require a maximum absolute voltage amplitude of 100 millivolts
    \item Require the presence of age, sex, diagnosis text, heart rate, PR interval, QT interval, QRS duration, P-axis, R-axis, and T-axis from each ECG
    \item P-axis must not be -1 degrees, because that was used as an indicator of missingness by the automated software
\end{enumerate}
We show summary statistics for BWH-20480 and BWH-test in Table \ref{tab:hospital_b_no_af}.
\begin{table}[htbp]
    \small
    \caption{BWH non-AF datasets}
    \vspace{1mm}
    \label{tab:hospital_b_no_af}
    \begin{minipage}{.5\linewidth}
        \centering
        \textbf{BWH-20480}\\
        \vspace{1mm}
            \begin{tabular}{lrrr}
            \toprule
            continuous feature &  median &   mean &    std \\
            \midrule
            age                  &   38.14 &  35.50 &  26.90 \\
            HR          &   72.00 &  75.19 &  17.07 \\
            PR            &  54.76 &  55.95 &  12.10 \\
            QRS           &   44.00 &  45.54 &   8.83 \\
            QT            &   89.26 &  89.30 &  11.52 \\
            R-axis &   32.00 &  29.90 &  41.49 \\
            T-axis &   42.00 &  44.81 &  39.41 \\
            P-axis &   52.00 &  50.02 &  23.66 \\
            \midrule
            binary feature & \% \\
            \midrule
            Female  &  55.49 \\
            Has LVH         &   3.97 \\
            Sample rate 250 Hz & \\
            (otherwise 500 Hz) &  46.41 \\
            \bottomrule
        \end{tabular}
    \end{minipage}%
    \begin{minipage}{.5\linewidth}
        \centering
        \textbf{BWH-test}\\
        \vspace{1mm}
        \begin{tabular}{rrr}
            \toprule
            median &   mean &    std \\
            \midrule
            39.71 &  38.00 &  27.39 \\
            73.00 &  75.84 &  17.84 \\
            54.76 &  56.09 &  12.57 \\
            44.00 &  46.30 &  10.14 \\
            89.26 &  89.47 &  12.11 \\
            31.00 &  29.21 &  44.92 \\
            42.00 &  46.83 &  44.82 \\
            51.00 &  48.63 &  25.39 \\
            \midrule
            \% \\
            \midrule
             54.13 \\
            3.95 \\
            \\
            47.21 \\
            \bottomrule
        \end{tabular}
    \end{minipage}
\end{table}

When AF is present in an ECG, the P-wave is missing, so the PR interval is not defined.
For that reason, we produce additional versions of each BWH dataset, AF-test with 10,000 ECGs, AF-640, AF-1280, etc.
Each of these datasets is a subset of the one larger than it.
For example, AF-1280 is a subset of AF-2560.
The selection criteria for ECGs in the AF datasets are the same as the criteria for the non-AF datasets, except we only require the presence of age, sex, heart rate, and diagnosis text from each ECG.
Summary statistics for AF-20480 and AF-test are shown in Table \ref{tab:hospital_b_af}.
\begin{table}[ht]
    \small
    \caption{BWH AF datasets}
    \vspace{1mm}
    \label{tab:hospital_b_af}
    \begin{minipage}{.5\linewidth}
        \centering
        \textbf{AF-20480}\\
        \vspace{1mm}
            \begin{tabular}{lrrr}
            \toprule
            continuous feature &  median &   mean &    std \\
            \midrule
            age         &   39.71 &  37.51 &  27.34 \\
            HR &   73.00 &  76.01 &  17.94 \\
            \midrule
            binary feature & \% \\
            \midrule
            Has AF &   5.27 \\
            Female &  54.79 \\
            Sample rate 250 Hz & \\ 
            (otherwise 500 Hz) &  46.59 \\
            \bottomrule
        \end{tabular}
    \end{minipage}%
    \begin{minipage}{.5\linewidth}
        \centering
        \textbf{AF-test}\\
        \vspace{1mm}
        \begin{tabular}{rrr}
            \toprule
            median &   mean &    std \\
            \midrule
            39.71 &  38.00 &  27.39 \\
            73.00 &  75.84 &  17.84 \\
            \midrule
            \% \\
            \midrule
            5.24 \\
            54.13 \\
            \\
            47.21 \\
            \bottomrule
        \end{tabular}
    \end{minipage}
\end{table}

When we use continuous features (e.g. age, PR interval) as inputs to linear models or as regression targets, we normalize by subtracting the mean of the training data and dividing by the standard deviation of the training data.

\section{Results}
\subsection*{When to use PCLR}
A critical question for clinical researchers is when to use PCLR rather than training a neural network from scratch.
In order to answer that question, we train PCLR on ECG representations extracted from BWH-640 through BWH-20480 to regress age and classify sex, LVH, and AF (AF on AF-640 through AF-20480).
All tasks besides AF detection are evaluated on BWH-test.
The ability to detect AF is evaluated on AF-test.
For comparison, we build and train a neural network with randomly initialized weights for each of the four tasks and six training dataset sizes.
To make the comparison fair, we use the same ECG encoder as we used to train PCLR (Figure \ref{fig:encoder}) followed by a linear fully connected layer.
In each classification task, the final layer outputs two values and the model is trained using the categorical cross entropy loss.
In age regression, the linear layer outputs one value and the models are trained using the mean squared error loss.
All of the models are trained using the Adam optimizer with learning rate $10^{-3}$ until the validation loss stops improving for five epochs.
For evaluation, we take the checkpoint of the network with the lowest validation loss.

\begin{figure}[h!]
  \centering
  \includegraphics[width={0.6 \textwidth}]{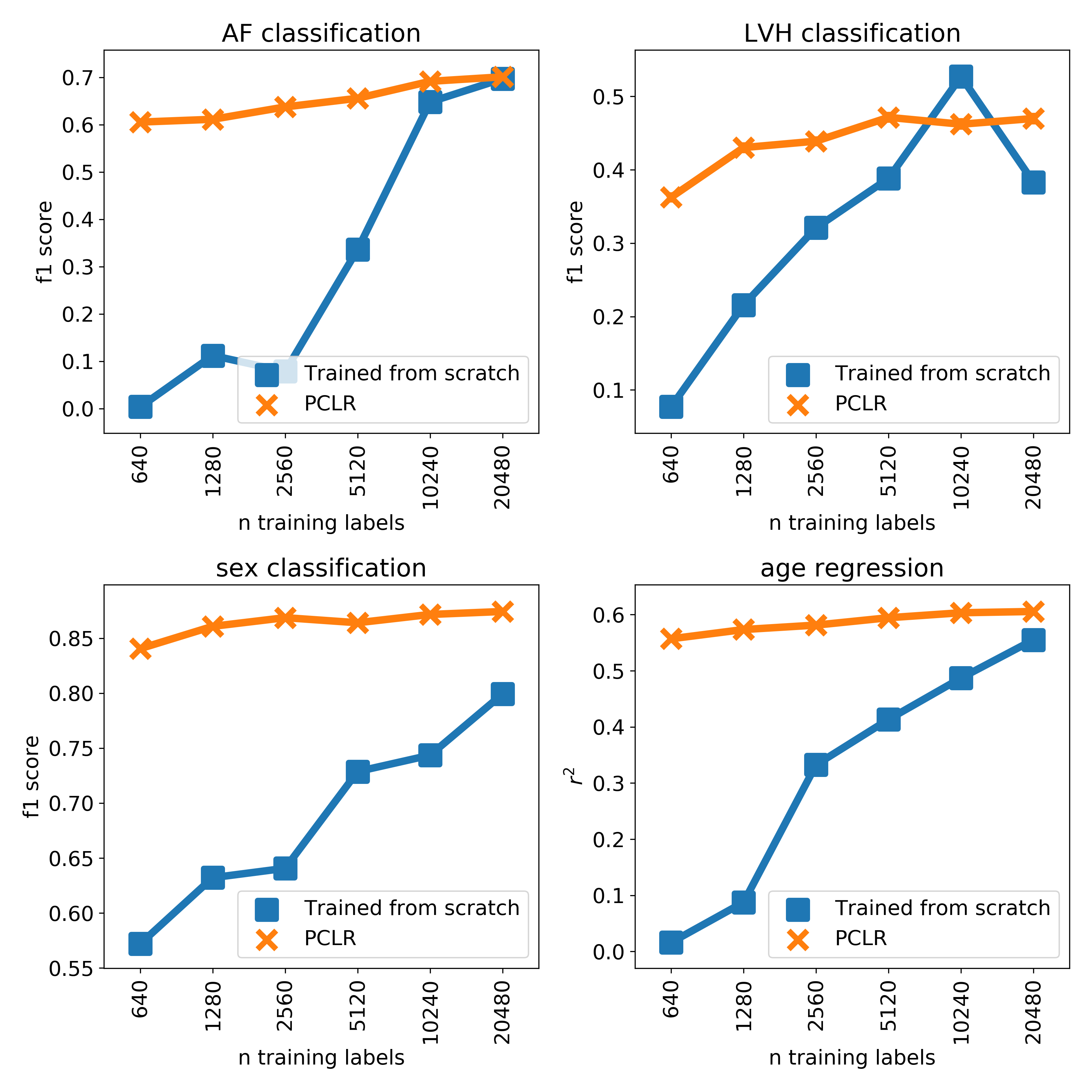}
  \label{fig:scratch_compare}
  \vspace{-6mm}
  \caption{Linear models trained on PCLR outperform training a neural network from scratch using thousands of labeled ECGs.}
\end{figure}

We find that PCLR is better than a model trained from scratch for all four tasks we evaluated up to at least 5,000 labeled training examples (Figure \ref{fig:scratch_compare}).
PCLR is especially effective at sex classification, which may be because sex is a patient specific property that, in our population, typically is same for all ECGs corresponding to a given patient. 

\vspace{-10mm}
\subsection*{Comparison to standard ECG features}
\begin{figure}[ht]
  \centering
  \includegraphics[width=.6\textwidth]{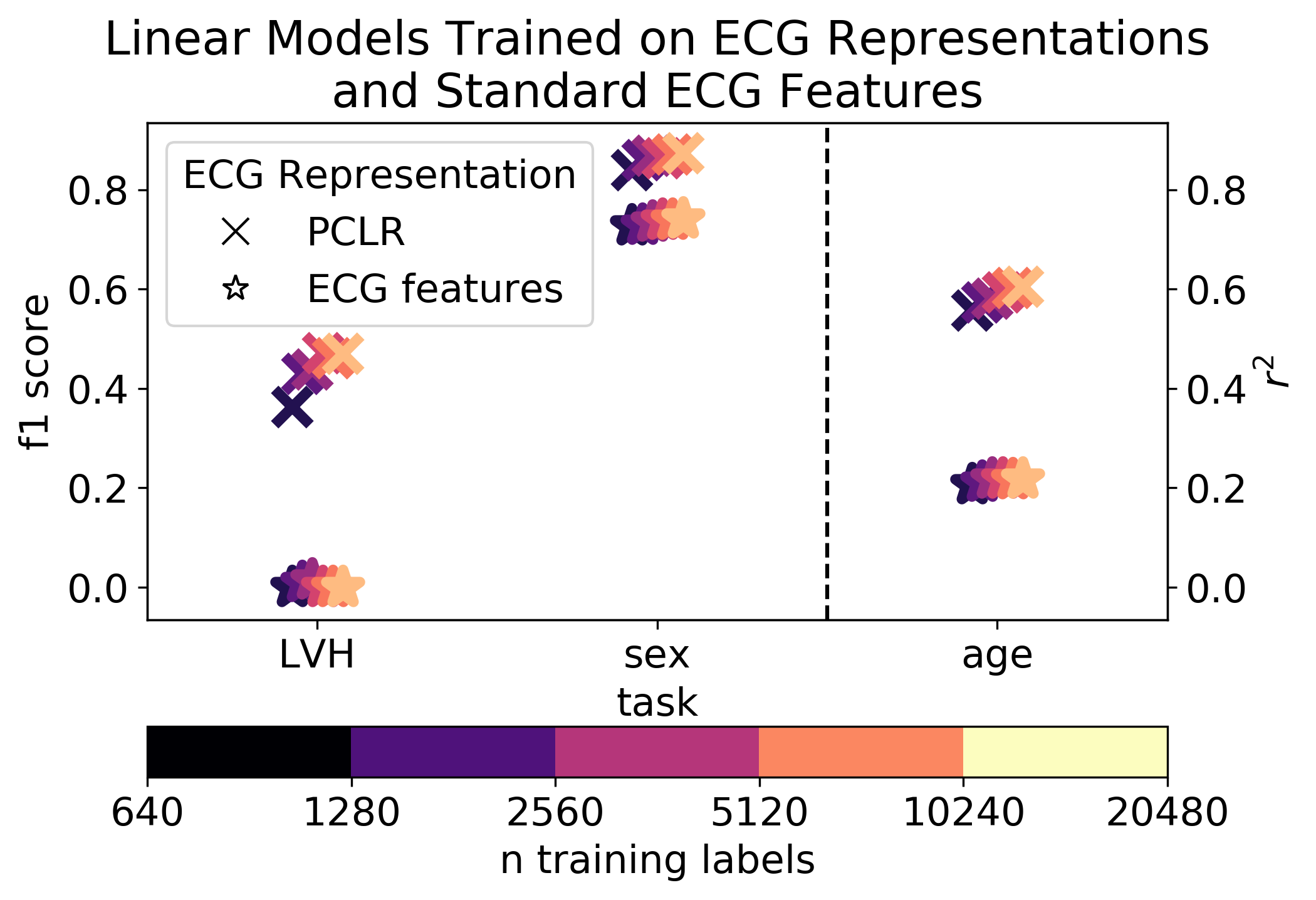} 
  \vspace*{-5mm}
  \caption{Comparison of linear models trained on PCLR to linear models trained on standard ECG features. Brighter color marks represent models trained on more examples.}
  \label{fig:ecg_feature_compare} 
\end{figure}

We compare using PCLR to using seven standard features extracted from the ECG: HR, PR interval, QRS duration, QT interval, P-axis, R-axis, and T-axis, which is another default approach for a clinician researcher.
The comparison is made in LVH and sex classification and age regression.
We cannot compare on AF classification, because some of the ECG features are undefined in AF. 
For this comparison, we train a linear model on the seven ECG features on all BWH training sets, and evaluate on the BWH test set.
Across all three tasks, PCLR is substantially better (Figure \ref{fig:ecg_feature_compare}).
These results show the advantage of using learned representations over hand designed features.
\vspace{-10mm}
\subsection*{Comparison to rhythm classification trained encoder }
\begin{figure}[ht]
    \centering
    \includegraphics[width=.62\textwidth]{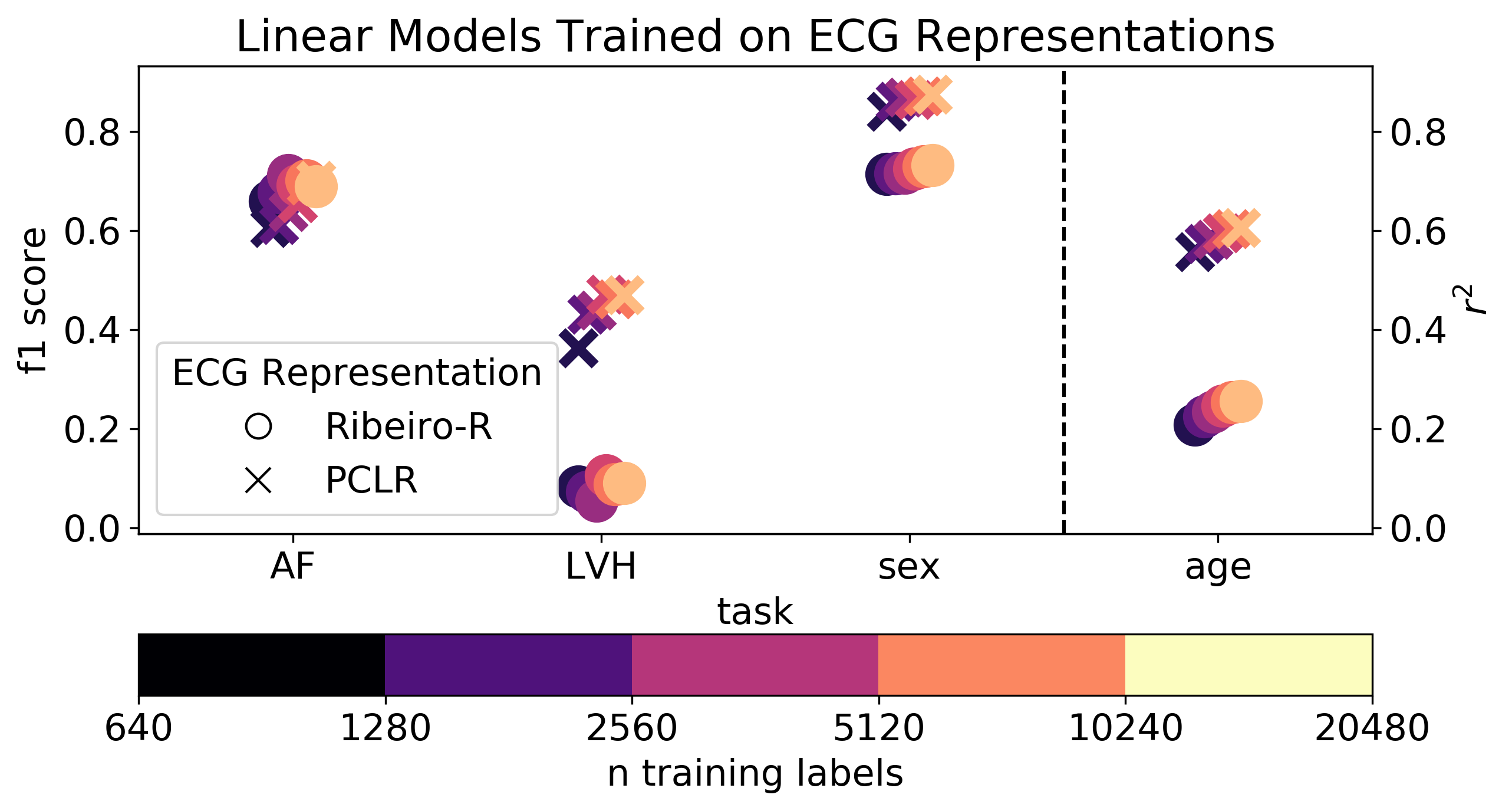}
    \vspace{-5mm}
    \caption{Comparison of the performance of linear models trained on PCLR and linear models trained ECG features extracted using a rhythm classification pre-trained encoder (Ribeiro-R). Brighter marks represent models trained on more examples.}
      \label{fig:picl_v_ribeiro}
\end{figure}

\cite{ribeiro} trained a model to classify 1st degree AV block, right bundle branch block, left bundle branch block, sinus bradycardia, atrial fibrillation, and sinus tachycardia using 2,322,513 ECG records from 1,676,384 different patients.
We apply linear evaluation (Section \ref{sec:lin_eval}) to the encoder portion of the Ribeiro model.
We term linear evaluation of ECG representations from the encoder with Ribeiro weights Ribeiro-R.
For this comparison, we train Ribeiro-R on all BWH training sets, and evaluate on the BWH test set.
Ribeiro-R is only competitive in AF classification  (Figure \ref{fig:picl_v_ribeiro}).
The particularly poor performance of Ribeiro-R in LVH classification could be because LVH is read from ECGs using voltage amplitudes \citep{cornell_lvh}, which may not be necessary to learn for rhythm classification.
\section{Discussion}
The application of deep learning to clinical datasets that have a few labeled training examples raises a number of issues.  
Fruitful application of neural networks to these data often requires the use of additional methods to mitigate the effects of small sample sizes.  
Consequently, the application of deep learning in the regime of limited training data in health care has required researchers with both in depth knowledge of machine learning and deep domain specific knowledge. Unfortunately, however, few researchers have both.

Here we develop a contrastive learning approach, PCLR, which helps resolve this gap. 
PCLR corresponds to a relatively low dimensional representation of high dimensional 12-lead ECG data.  A key aspect of the approach is that these representations are constructed in a manner to ensure that different ECGs, which arise from the same patient, are more similar to one another relative to ECG-representations arising from different patients.  
This is a unique-to-health-care self-supervised approach for building useful representation of rich medical data. We demonstrate that PCLR is performant, expressive, and practical for clinical researchers to use across a variety of tasks. We also outline the regimes where PCLR is the most performant approach to adopt.

The success of PCLR suggests that building patient-centered representations of multi-modal health care data, not just ECGs, is an important direction of future research.  

\acks{
    Many thanks to Daphne Schlesinger, Aniruddh Raghu, Samuel Friedman, and Marcus Klarqvist for useful discussion throughout the research process.
    Thank you, Paolo Di Achille for your work preparing ECG data for our team to use.
    Thank you Shaan Khurshid, 
    Emily Lau, Jennifer Ho, Steven Lubitz, and Anthony Philippakis for supporting the research process throughout.
    Thank you to IBM, Bayer AG and Quanta Inc whose funding helped support this work.
}

\bibliography{main}
\newpage
\appendix
\section*{Appendix A: PCLR pre-training Architecture}
All activations are ReLU, and all convolutions have filter size of 16.
\begin{lstlisting}[basicstyle=\tiny, label={lst:model}]
__________________________________________________________________________________________________
Layer (type)                    Output Shape         Param #     Connected to
==================================================================================================
ecg (InputLayer)                [(None, 4096, 12)]   0
__________________________________________________________________________________________________
conv1d (Conv1D)                 (None, 4096, 64)     12288       ecg[0][0]
__________________________________________________________________________________________________
batch_normalization (BatchNorma (None, 4096, 64)     256         conv1d[0][0]
__________________________________________________________________________________________________
activation (Activation)         (None, 4096, 64)     0           batch_normalization[0][0]
__________________________________________________________________________________________________
conv1d_2 (Conv1D)               (None, 4096, 128)    131072      activation[0][0]
__________________________________________________________________________________________________
batch_normalization_1 (BatchNor (None, 4096, 128)    512         conv1d_2[0][0]
__________________________________________________________________________________________________
activation_1 (Activation)       (None, 4096, 128)    0           batch_normalization_1[0][0]
__________________________________________________________________________________________________
max_pooling1d (MaxPooling1D)    (None, 1024, 64)     0           activation[0][0]
__________________________________________________________________________________________________
conv1d_3 (Conv1D)               (None, 1024, 128)    262144      activation_1[0][0]
__________________________________________________________________________________________________
conv1d_1 (Conv1D)               (None, 1024, 128)    8192        max_pooling1d[0][0]
__________________________________________________________________________________________________
add (Add)                       (None, 1024, 128)    0           conv1d_3[0][0]
                                                                 conv1d_1[0][0]
__________________________________________________________________________________________________
batch_normalization_2 (BatchNor (None, 1024, 128)    512         add[0][0]
__________________________________________________________________________________________________
activation_2 (Activation)       (None, 1024, 128)    0           batch_normalization_2[0][0]
__________________________________________________________________________________________________
conv1d_5 (Conv1D)               (None, 1024, 196)    401408      activation_2[0][0]
__________________________________________________________________________________________________
batch_normalization_3 (BatchNor (None, 1024, 196)    784         conv1d_5[0][0]
__________________________________________________________________________________________________
activation_3 (Activation)       (None, 1024, 196)    0           batch_normalization_3[0][0]
__________________________________________________________________________________________________
max_pooling1d_1 (MaxPooling1D)  (None, 256, 128)     0           add[0][0]
__________________________________________________________________________________________________
conv1d_6 (Conv1D)               (None, 256, 196)     614656      activation_3[0][0]
__________________________________________________________________________________________________
conv1d_4 (Conv1D)               (None, 256, 196)     25088       max_pooling1d_1[0][0]
__________________________________________________________________________________________________
add_1 (Add)                     (None, 256, 196)     0           conv1d_6[0][0]
                                                                 conv1d_4[0][0]
__________________________________________________________________________________________________
batch_normalization_4 (BatchNor (None, 256, 196)     784         add_1[0][0]
__________________________________________________________________________________________________
activation_4 (Activation)       (None, 256, 196)     0           batch_normalization_4[0][0]
__________________________________________________________________________________________________
conv1d_8 (Conv1D)               (None, 256, 256)     802816      activation_4[0][0]
__________________________________________________________________________________________________
batch_normalization_5 (BatchNor (None, 256, 256)     1024        conv1d_8[0][0]
__________________________________________________________________________________________________
activation_5 (Activation)       (None, 256, 256)     0           batch_normalization_5[0][0]
__________________________________________________________________________________________________
max_pooling1d_2 (MaxPooling1D)  (None, 64, 196)      0           add_1[0][0]
__________________________________________________________________________________________________
conv1d_9 (Conv1D)               (None, 64, 256)      1048576     activation_5[0][0]
__________________________________________________________________________________________________
conv1d_7 (Conv1D)               (None, 64, 256)      50176       max_pooling1d_2[0][0]
__________________________________________________________________________________________________
add_2 (Add)                     (None, 64, 256)      0           conv1d_9[0][0]
                                                                 conv1d_7[0][0]
__________________________________________________________________________________________________
batch_normalization_6 (BatchNor (None, 64, 256)      1024        add_2[0][0]
__________________________________________________________________________________________________
activation_6 (Activation)       (None, 64, 256)      0           batch_normalization_6[0][0]
__________________________________________________________________________________________________
conv1d_11 (Conv1D)              (None, 64, 320)      1310720     activation_6[0][0]
__________________________________________________________________________________________________
batch_normalization_7 (BatchNor (None, 64, 320)      1280        conv1d_11[0][0]
__________________________________________________________________________________________________
activation_7 (Activation)       (None, 64, 320)      0           batch_normalization_7[0][0]
__________________________________________________________________________________________________
max_pooling1d_3 (MaxPooling1D)  (None, 16, 256)      0           add_2[0][0]
__________________________________________________________________________________________________
conv1d_12 (Conv1D)              (None, 16, 320)      1638400     activation_7[0][0]
__________________________________________________________________________________________________
conv1d_10 (Conv1D)              (None, 16, 320)      81920       max_pooling1d_3[0][0]
__________________________________________________________________________________________________
add_3 (Add)                     (None, 16, 320)      0           conv1d_12[0][0]
                                                                 conv1d_10[0][0]
__________________________________________________________________________________________________
batch_normalization_8 (BatchNor (None, 16, 320)      1280        add_3[0][0]
__________________________________________________________________________________________________
activation_8 (Activation)       (None, 16, 320)      0           batch_normalization_8[0][0]
__________________________________________________________________________________________________
embed (GlobalAveragePooling1D)  (None, 320)          0           activation_8[0][0]
__________________________________________________________________________________________________
projection_0 (Dense)            (None, 320)          102720      embed[0][0]
__________________________________________________________________________________________________
projection (Dense)              (None, 320)          102720      projection_0[0][0]
==================================================================================================
Total params: 6,600,352
Trainable params: 6,596,624
Non-trainable params: 3,728
\end{lstlisting}

\section*{Appendix B: PCLR minibatch pre-training procedure}

\begin{lstlisting}[language=Python]
def build_PICL_batch(batch_of_patients):
    """
    param batch_of_patients: 
    A list of collections of ECGs per patient
    
    returns: 
    A list of ECGs to use as input to a PCLR model
    of length twice the number of patients given
    """
    first_half_batch = []
    second_half_batch = []
    for i in range(len(batch_of_patients)):
        patient_i_first_ecg = random.choice(batch_of_patients[i])
        patient_i_second_ecg = random.choice(batch_of_patients[i])
        # note that we choose with replacement so
        # sometimes patient_i_first_ecg == patient_i_second_ecg
        
        first_half_batch.append(patient_i_first_ecg)
        second_half_batch.append(patient_i_second_ecg)
        
    # concatenate and return the halves of the batch
    return first_half_batch + second_half_batch
\end{lstlisting}

\section*{Appendix C: AF and LVH diagnosis keywords}\label{app:keywords}
\begin{lstlisting}[label={lst:keywords}, language=Python]
af_key_words = {
    "atrial fibrillation with rapid ventricular response",
    "atrial fibrillation with moderate ventricular response",
    "fibrillation/flutter",
    "atrial fibrillation with controlled ventricular response",
    "afib",
    "atrial fib",
    "afibrillation",
    "atrial fibrillation",
    "atrialfibrillation",
}
lvh_keywords = {
    "biventricular hypertrophy",
    "leftventricular hypertrophy",
    "combined ventricular hypertrophy",
    "left ventricular hypertr",
    "biventriclar hypertrophy",
}


def is_af(diagnosis_text: str):
    for keyword in af_key_words:
        if keyword in diagnosis_text.lower():
            return True
    return False


def is_lvh(diagnosis_text: str):
    for keyword in lvh_key_words:
        if keyword in diagnosis_text.lower():
            return True
    return False
\end{lstlisting}

\end{document}